\documentclass[10pt,twocolumn,letterpaper]{article}

\usepackage{cvpr}              

\usepackage{graphicx}
\usepackage{amsmath}
\usepackage{amssymb}
\usepackage{booktabs}
\usepackage{pifont}%
\usepackage{adjustbox}
\usepackage{multirow}

\usepackage{caption}
\usepackage{subcaption}
\usepackage{comment}
\usepackage[table]{xcolor}  
\usepackage{xcolor}
\usepackage{soul}
\usepackage{tabularx} 

\usepackage[pagebackref,breaklinks,colorlinks]{hyperref}
\usepackage[accsupp]{axessibility}  

\usepackage[normalem]{ulem}

\usepackage[capitalize]{cleveref}
\crefname{section}{Sec.}{Secs.}
\Crefname{section}{Section}{Sections}
\Crefname{table}{Table}{Tables}
\crefname{table}{Tab.}{Tabs.}

\usepackage{color, colortbl}
\definecolor{arthurcolor}{RGB}{0, 60, 160}

\def\cvprPaperID{7895} 
\def\confName{CVPR}
\def\confYear{2022}

\begin{document}

\title{What to look at and where: Semantic and Spatial Refined Transformer \\ for detecting human-object interactions}

\author{A S M Iftekhar\thanks{Equal contribution.} \thanks{Work done during an internship at Amazon.}, $\;$ Hao Chen\footnotemark[1] \textsuperscript{\rm $\ddagger$}, $\;$ Kaustav Kundu\textsuperscript{\rm $\ddagger$}, $\;$ Xinyu Li\textsuperscript{\rm $\ddagger$}, $\;$ Joseph Tighe\textsuperscript{\rm $\ddagger$}, $\;$Davide Modolo\textsuperscript{\rm $\ddagger$}\\
\textsuperscript{\rm $\ddagger$}AWS AI Labs; \textsuperscript{\rm $\dagger$}University of California, Santa Barbara\\
{\tt\small \{hxen,kaustavk,-,tighej,dmodolo\}@amazon.com; iftekhar@ucsb.edu}
}

\maketitle

\begin{abstract}

We propose a novel one-stage Transformer-based semantic and spatial refined transformer (SSRT) to solve the Human-Object Interaction detection task, which requires to localize humans and objects, and predicts their interactions. Differently from previous Transformer-based HOI approaches, which mostly focus at improving the design of the decoder outputs for the final detection, SSRT introduces two new modules to help select the most relevant object-action pairs within an image and refine the queries' representation using rich semantic and spatial features. These enhancements lead to state-of-the-art results on the two most popular HOI benchmarks: V-COCO and HICO-DET.

\vspace{-0.2cm}
 \end{abstract}

\section{Introduction}

\begin{figure}[t]
\begin{center}
\includegraphics[width=0.95\linewidth]{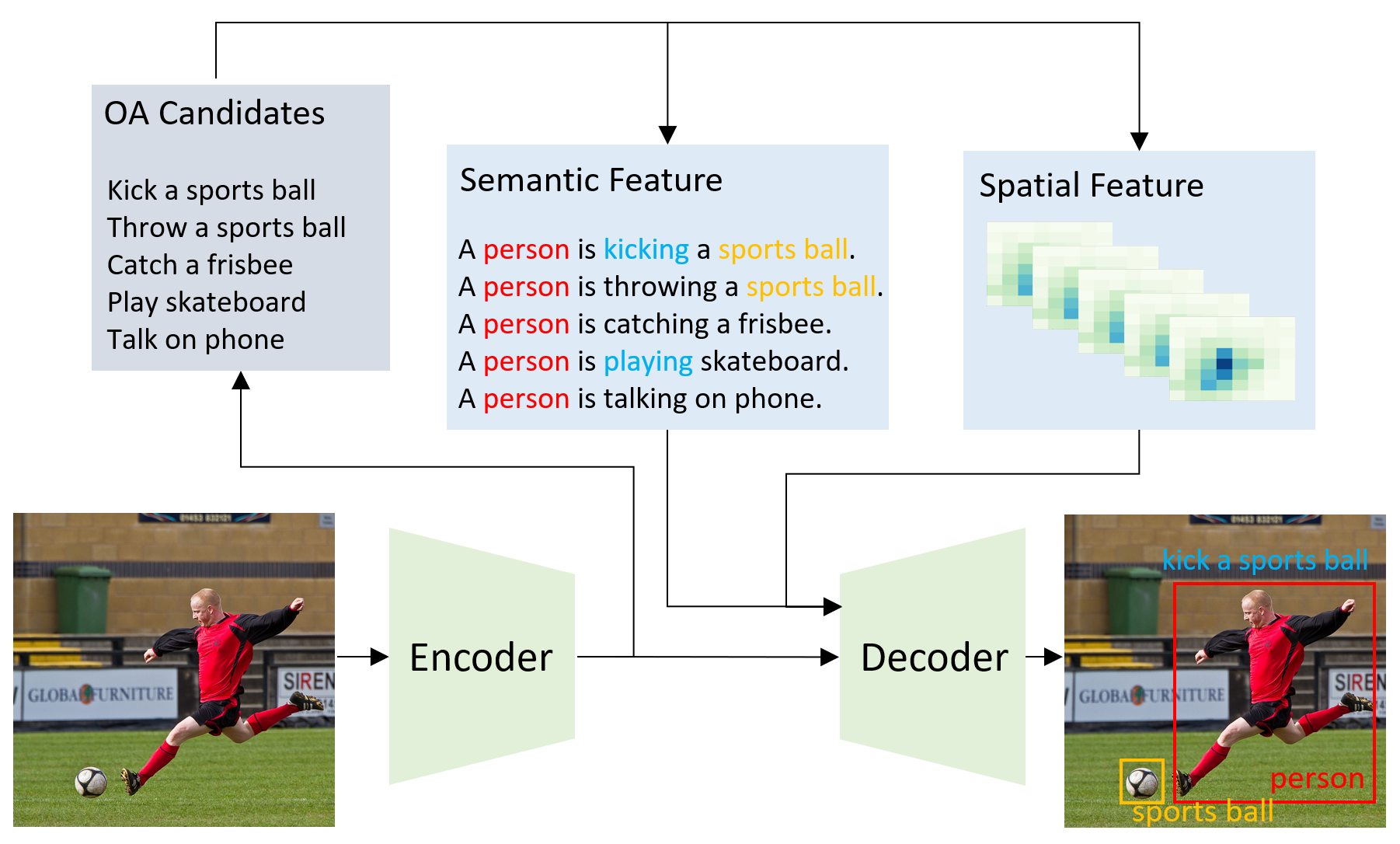}
\end{center}
  \caption{\small \it Conceptual workflow of SSRT.
  Instead of just feeding the encoded image to the decoder, we pre-select {\it object-action (OA)} prediction candidates and encode them to semantic and spatial features. These features then refine the learnt queries in decoding to enable them to attend to more relevant HOI predictions.}
\label{Fig:intro}
\vspace{-0.6cm}
\end{figure}

Human-object interaction (HOI) detection is an important building block for complex visual reasoning, such as scene understanding~\cite{eslami2016attend,xiao2018unified} and action recognition~\cite{wu2019learning,zhang2019structured}, and its goal is to detect all HOI triplets $\langle human, object, action \rangle$ in each image. Fig.~\ref{Fig:intro} shows an example of a HOI detection, where the {\it person} (i.e., the human) is denoted with a red bounding box, the {\it sports ball} (i.e., the object) with a yellow bounding box, and the action {\it kick} is what that human is performing with that object.

The HOI literature can be divided into \emph{two-stage} and \emph{one stage} approaches.
\emph{Two-stage} approaches~\cite{fang2020dirv,gao2018ican,li2019transferable,liu2020amplifying, wan2019pose, wang2019deep,gao2020drg,ulutan2020vsgnet, qi2018learning, zhong2020polysemy, kim2020detecting, liu2020consnet, hou2020visual, wang2020contextual, IDN, he2021exploiting,zhang2021spatially,yang2021rr} first use off-the-shelf detectors to localize all instances of people and objects independently. For each person and object bounding box pair, an interaction class is then predicted in the second stage.
This sequential process has two main drawbacks~\cite{chen2021reformulating, kim2021hotr, kim2020uniondet}: (1) off-the-shelf object detectors are agnostic to the concept of interactions; and (2) enumerating over all pairs of person and object bounding boxes to predict an interaction class is time-consuming and expensive.
In contrast, \emph{one-stage} approaches detect all the components of an HOI triplet directly in an end-to-end fashion. Some earlier one-stage approaches used intermediate representations based on interaction points~\cite{liao2020ppdm,wang2020learning} and union boxes~\cite{kim2020uniondet} to predict these. However, such methods fail when the interacting human and object are far away from each other and when multiple interactions overlap (e.g., crowd scenes)~\cite{tamura2021qpic,chen2021reformulating}. 

More recently, a new trend of one-stage approaches~\cite{tamura2021qpic, chen2021reformulating, kim2021hotr, zou2021end} based on \emph{Transformer architectures}~\cite{carion2020end,dosovitskiy2020image,liu2021swin} have been proposed to overcome these problems and improve the HOI detection performance. This paper belongs to this category of works (Fig.~\ref{Fig:intro}). 
At a high-level, these approaches first use a CNN backbone to extract image features and then feed them into an encoder-decoder architecture. Some approaches use two decoders to detect instances and interactions in parallel~\cite{chen2021reformulating, kim2021hotr}. while others follow a simpler design that directly predicts all the elements of an HOI triplet with a single decoder~\cite{zou2021end,tamura2021qpic}. While successful, this design suffers from two limitations: (i) not all object-action pairs are meaningful (e.g., a person cannot be `cutting a pizza' when the pizza is far away from the person's location; and it is unusual for a person to be `cutting a football'), simply relying on the one-shot network to reduce them may not be effective; and (ii) each simple query is decoded for all the rich elements of an HOI triplet (i.e., person location, object class and location, and interaction class), which is challenging, especially considering how HOI detection requires reasoning about complex relational structures in images. We address both of these limitations in our work.

\begin{figure*}[t]
\begin{center}
\includegraphics[width=\linewidth]{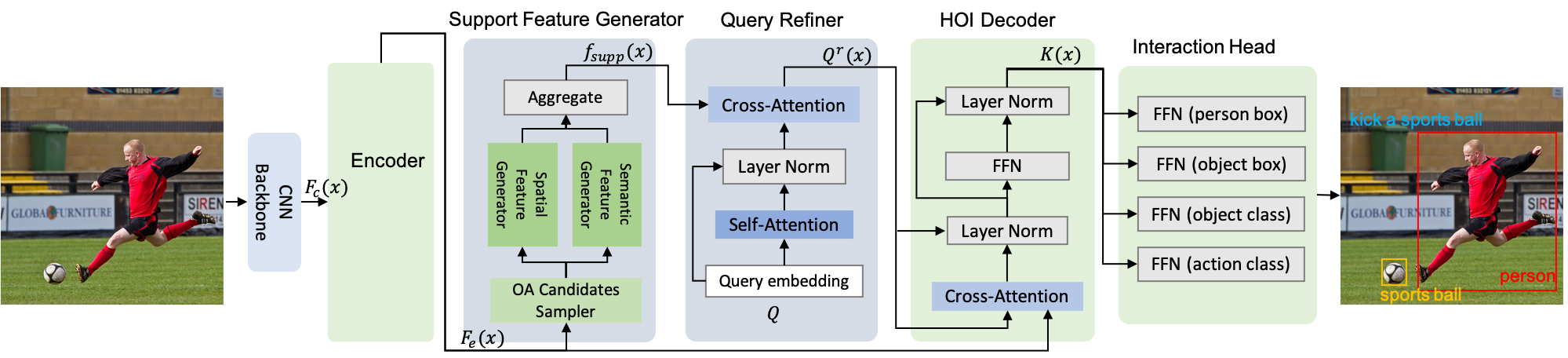}
\end{center}
\vspace{-0.4cm}
   \caption{\small \it Overview of our SSRT network. Given an image, we extract features using a backbone and feed them to Transformer Encoder. The encoder's output is then sent to the Support Feature Generator (SFG), which first predicts and selects top-K object-action (OA) candidates, and then generates spatial and semantic features. Next, the aggregated features are sent to the Query Refiner (QR) to refine queries. Finally, The refined queries are decoded: each query is used to predict a human bounding box, an object bounding box, an interaction vector and an object class vector via interaction heads with small FFNs. 
   }
\label{fig:model_architecture}
\vspace{-0.2cm}
\end{figure*}

For this, we propose {\bf S}emantic and {\bf S}patial {\bf R}efined {\bf T}ransformer (SSRT), that solves the aforementioned limitations by predicting what subset of object-action pairs is relevant for an image, and using explicit semantic and spatial information to support and guide the queries, so that they can be decoded more reliably and are more aligned with the final detection. In details, SSRT improves the Transformer design of previous HOI detector by introducing two novel modules: a Support Feature Generator (SFG) and a Query Refiner (QR) (Fig.~\ref{fig:model_architecture}). The former generates semantic and spatial features from a set of pre-selected {\it object-action (OA)} pairs, while the latter integrates these features for decoding. 

Our approach achieves state-of-the-art results on both V-COCO ~\cite{gupta2015visual} and HICO-DET~\cite{chao2018learning} datasets, showing the importance and effectiveness of our semantic and spatial guidance for HOI detection. Finally, in an extensive ablation study, we also evaluate our model design and our parameter choices, to further highlight the SSRT's contributions.

\section{Related Work}

Two stage HOI detection networks detect objects and then detect HOIs among those detected objects. These networks rely on an off-the-shelf object detector~\cite{ren2015faster} to localize objects. For detecting interactions among the detected objects these networks develop different novel techniques. Few works ~\cite{gao2020drg, liu2020consnet} consider humans and objects as nodes in graph networks. Another line of works~\cite{ulutan2020vsgnet, wan2019pose} utilize spatial and pose features to attend salient spatial regions of the images. Additionally, other works are using object affordance based architectures~\cite{hou2021affordance, hou2021detecting} to deal with the long-tail distribution problem of the HOI detection datasets. Moreover, there are works~\cite{IDN,hou2020visual} that leverage the compositional nature of objects and interactions to detect HOIs. Another paradigm of two stage works utilize additional features like 3D representation of humans~\cite{li2020detailed}, semantic contexts~\cite{iftekhar2021gtnet, liu2020consnet}, segmentation masks~\cite{liu2020amplifying}. However, performance of these networks are highly dependent on the quality of the object detection. Moreover, these networks suffer heavily to process the overwhelming number of non-interacting detected objects~\cite{zhong2021glance}. 

To deal with the issues faced by two stage networks, recent works~\cite{tamura2021qpic, chen2021reformulating, kim2021hotr, zou2021end, zhong2021glance, dong2021visual} are trying to detect HOIs in a one stage framework. These networks take images as input and directly detect and localize HOIs over those images. Initial one stage HOI detection networks~\cite{liao2020ppdm, wang2020learning} focus on detecting pre-defined interaction points to detect interactions. However, these heuristic based approaches often fail to find spatial contextual information. For getting richer contextual features many recent one stage HOI detection networks~\cite{tamura2021qpic, chen2021reformulating, kim2021hotr, zou2021end} adapt  encoder-decoder based Transformer~\cite{vaswani2017attention} like architecture inspired from the one stage object detection network DETR~\cite{carion2020end}.

However, these networks do not consider the additional complexity of doing two related but different subtasks of object localization and interaction detection. The base network of these mentioned works is essentially an object detector network which is expanded for interaction detection. Therefore, it is beneficial to provide additional guidance to these networks. Moreover, these one stage networks do not leverage spatial and semantic cues that are proven to be beneficial to detect HOIs in few two-stage works~\cite{ulutan2020actor, xu2019learning, iftekhar2021gtnet}. In this respect, we propose a semantic and spatially refined transformer based architecture to detect HOIs in one single stage. Our superior numerical results over these state of the art methods prove our method's effectiveness.

\vspace{-1.5mm}
\section{Technical Approach}

Most of today's Transformer-based HOI detection networks~\cite{tamura2021qpic, chen2021reformulating, kim2021hotr, zou2021end} follow the DETR~\cite{carion2020end} architecture and focus on improving the \emph{design of the decoder outputs} for the HOI task. Instead, our SSRT approach improves the overall \emph{design of the Transformer}. Specifically, it adds two new modules between the encoder and the decoder: a Support Feature Generator ({\bf SFG}) (Sec.~\ref{subsec:sem&spa}) and a Query Refiner ({\bf QR}) (Sec.~\ref{subsec:cr}) (Fig.~\ref{fig:model_architecture}).
At a high level, SSRT works as follows: given an input image, it first extracts its features with a CNN backbone and then transform those using a transformer encoder. Instead of feeding the encoded features directly into the decoder, the features are sent to SFG to first generate a set of {\it object-action (OA)} prediction candidates (without localization). Then the SFG generates both spatial and semantic features using these candidates and aggregates them as support features. These support features are then sent to the QR to refine the learnable queries. Finally, the HOI Decoder takes the inputs as both the encoded features and the refined queries, and outputs a set of embeddings, each of which is used to predict an HOI output.

\subsection{Our Architecture}\label{subsec:ov}
 Given an input image ${\bf x} \in \mathbb{R}^{H_0\times W_0\times C_0}$, where $H_0,W_0,C_0$ denote the image height, width and color channels, SSRT first extracts a feature map $\mathbb{R}^{H\times W
\times C}$ using a CNN backbone ($ {\bf F}  $) (e.g. ResNet-50~\cite{he2016deep}). 
${\bf F} ({\bf x}) $ is then sent to a 1x1 convolution to reduce the channel dimension $C$ to a smaller value $d$, and obtain ${\bf F}_c ({\bf x}) \in \mathbb{R} ^{H \times W \times d}$. Following previous works~\cite{tamura2021qpic, chen2021reformulating, kim2021hotr, zou2021end}, we add a fixed positional encoding ${\bf p} \in \mathbb{R} ^{H \times W \times d}$ to the input feature of the encoder to supplement the positional information.
The encoder follows the standard architecture of the transformer as a stack of multi-head self-attention modules and a feed forward network (FFN). The encoded feature map ${\bf F}_e ({\bf x}) \in \mathbb{R} ^{H \times W \times d}$ is obtained as follows:
\vspace{-1.5mm}
\begin{equation}
\begin{small}
\vspace{-1.5mm}
        {\bf F}_e ({\bf x}) = \text{Encoder}({\bf F}_c ({\bf x}), {\bf p} )
\end{small}        
\end{equation}

Instead of feeding $ {\bf F}_e ({\bf x}) $ only to the HOI decoder, we also send it to our SFG module. Here, we first predict $ K $ pairs of (object, action) categories (i.e., OA pairs) present in the image and select a subset from them. We then use spatial and semantic cues for the OA prediction candidates to generate support features, $ {\bf f}_{supp}({\bf x}) $. 
The details of the SFG module are discussed in Sec.~\ref{subsec:sem&spa}.
The support features, $ {\bf f}_{supp}({\bf x}) $ along with an initial set of queries, ${\bf Q} = \{ {\bf q_i} | {\bf q_i} \in \mathbb{R}^d \}_{i=1}^{N_q}$ are fed into the QR module. The query refiner module is a decoder like architecture which outputs a refined set of queries, ${\bf Q}^r({\bf x}) = \{ {\bf q_i^r({\bf x})} | {\bf q_i^r({\bf x})} \in \mathbb{R}^d \}_{i=1}^{N_q}$, i.e., %
\begin{equation}
\begin{small}
        {\bf Q}^r({\bf x}) = \text{QR}({\bf f}_{supp}({\bf x}), {\bf Q})
\end{small}
\end{equation} 
The details of the QR block are discussed in Sec.~\ref{subsec:cr}. Both the encoder output ${\bf F}_e ({\bf x})$ and the refined queries ${\bf Q}^r({\bf x}) $ are sent to the HOI decoder for the final decoding. Note that in contrast to the standard transformer architectures, the queries which are fed into the decoder are a function of the input image ${\bf x}$. The goal of modeling it in this manner is to explicitly provide more guidance to the decoder so that it can generate more accurate HOI outputs. The HOI decoder follows the standard architecture of the transformer as a stack of multi-headed cross-attention units but no self-attention layers. The refined queries ${\bf Q}^r({\bf x})$ are transformed into a set of output embeddings, ${\bf K}({\bf x}) = \{ {\bf k}_i({\bf x}) | {\bf k}_i({\bf x}) \in \mathbb{R}^d \}_{i=1}^{N_q}$, i.e.,:
\begin{equation}
\vspace{-1.5mm}
\begin{small} 
        {\bf K}({\bf x}) = \text{Decoder}({\bf F}_e ({\bf x}), {\bf p} , {\bf Q}^r({\bf x}))
\end{small} 
\end{equation} 
where ${\bf p}$ is the positional embedding.
Each query is designed to capture at most one HOI prediction. We feed these queries to four small Feed Forward Network (FFNs) to predict human bounding boxes ${\bf b}_{h}({\bf x}) \in [0,1]^4$, object bounding boxes ${\bf b}_{o}({\bf x}) \in [0, 1]^4$, interaction prediction vectors ${\bf P}_{HOI} ({\bf x}) \in  [0,1]^{N_{act}}$, and object class prediction vectors ${\bf P}_{obj}({\bf x}) \in [0,1]^{N_{obj}}$, where $N_{act}$ and $N_{obj}$ are the number of interaction classes and number of object classes. ${\bf b}_{o}$, ${\bf b}_{h}$ and ${\bf P}_{HOI}$ are predicted with sigmoid functions, ${\bf P}_{obj}$ is predicted with softmax function. Like~\cite{tamura2021qpic}, we weigh our final interaction prediction vectors with the most confident object class predictions as: 
\begin{equation}
\vspace{-1.5mm}
\begin{small} 
     {\bf P}_{HOI}({\bf x})= {\bf P}_{HOI}({\bf x}) * max({\bf P}_{obj}({\bf x})) 
\end{small} 
\end{equation} 
We discuss details on how to train this network in Sec.~\ref{sec:train}.

\begin{figure}[t]
\begin{center}
\includegraphics[width=0.95\linewidth]{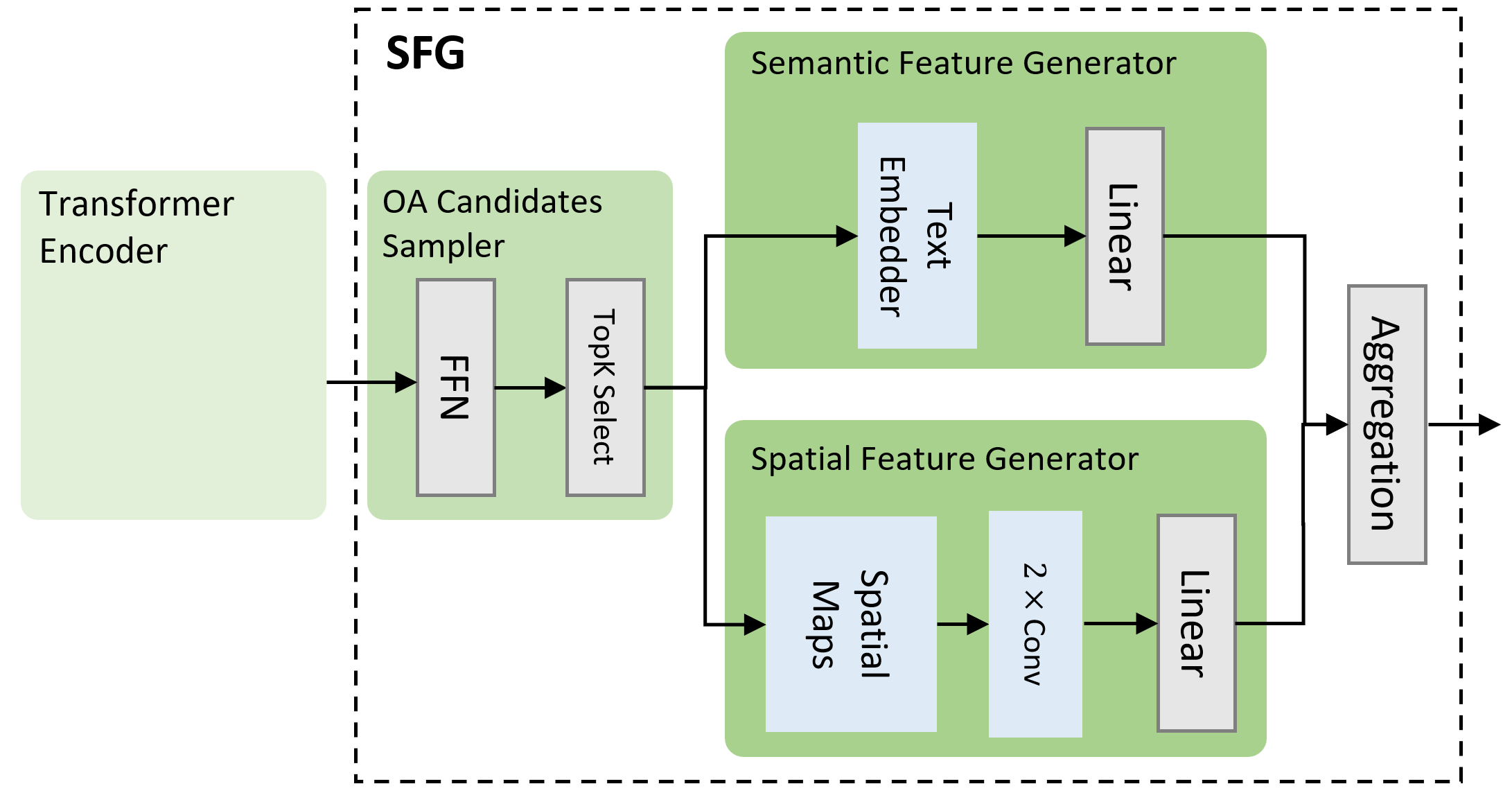}
\end{center}
\vspace{-0.4cm}
   \caption{ \small \it Our Support Feature Generator (SFG) design. 
   }
\label{fig:model_architecture_sem_spa}
\end{figure} 

\subsection{Support Feature Generator}\label{subsec:sem&spa}

The goal of SFG is to provide support to the transformer with additional semantic and spatial cues, as they play significant roles towards detecting all the rich HOI outputs. Specifically, semantic cues are important to help capture the human-object relations~\cite{xu2019learning} while spatial cues are important to help accurately localize the humans and objects~\cite{ulutan2020vsgnet}. 

While explicitly using these cues have been proved successful in the two-stage solutions~\cite{xu2019learning,ulutan2020vsgnet}, it has not been exploited in one-stage approaches. In this block, we propose how to generate support features  $  {\bf f}_{supp}({\bf x})  $ for an image, which can then be used as inputs to the query refiner block and subsequently to the decoder.  
We first select $ K $ high confident {\it object-action (OA)} candidates predicted from the encoder feature, $ {\bf F}_e ({\bf x}) $, and enrich them with semantic and spatial embeddings to generate the support features (Fig.~\ref{fig:model_architecture_sem_spa}).

\vspace{-3mm}
\paragraph{OA Candidates Sampler.}
As shown in Fig.~\ref{fig:model_architecture_sem_spa}, we build a 3-layer FFN $g_{cls}$ on top of the average-pooled encoded features ${\bf F}_e ({\bf x})$ to predict the {\it object-action (OA)} candidates, i.e., ${\bf s} ({\bf x}) = \sigma(g_{cls}( \text{avg-pool}({\bf F}_e ({\bf x}))))$, where ${\bf s} ({\bf x}) \in [0,1]^{N_s}$, $N_s$ is the number of possible set of object-action (OA) pairs and $\sigma$ is the sigmoid function. Note how in this module, ${\bf s} ({\bf x})$ corresponds to the OA labels without localization. 
Among all predictions, we then select the top-K (object, action) candidates with the highest confidence. Let this set of selected candidates be represented by $ {\bf S}_{\text{cand}} = \{ (y_{o, i}, y_{a, i}) \}_{i = 1}^K $. 

\vspace{-3mm}
\paragraph{Semantic Feature Generator.} 
Recently, language-image models~\cite{radford2021learning,jia2021scaling} have shown strong capabilities in generating high quality representations that can capture rich semantic and context information. We believe such embedding should be able to capture the relational structure and enrich the context of the transformer network.  Therefore, we use a CLIP~\cite{radford2021learning} text encoder to compute the semantic representation of each pre-selected OA candidate. Since CLIP works best with sentences (as opposed to single words), we convert each predicted OA into a full sentence before feeding it to the CLIP text encoder. For example, we transform the pair $(phone, talk)$ into the sentence ``A person is talking on the phone''. The transformation is done automatically following certain pre-defined rules using scripts with tiny manual efforts. Finally we project these semantic features to the image feature space by using a linear projection layer (Fig.~\ref{fig:model_architecture_sem_spa}, top). For each OA candidate, $ (y_{o, i}, y_{a, i}) \in {\bf S}_{\text{cand}} $, we compute the semantic feature as follows:
\begin{equation}
\begin{small}
        {\bf f}_{sem}(y_{o, i}, y_{a, i})= \text{Emb}_{sem}(y_{o, i}, y_{a, i})
\end{small} 
\end{equation} 
where, $\text{Emb}_{sem}$ is the semantic embedding function. 

\vspace{-3mm}
\paragraph{Spatial Feature Generator.}
In two-stage HOI approaches, the spatial features are generated based on predicted human and object bounding boxes from off-the-shelf detectors~\cite{ulutan2020vsgnet, gao2018ican}. As these are not available for one-state approaches like ours, we propose to estimate bounding box locations using training data statistics.

We define the {\it relative spatial configuration (RSC)} as the object bounding box location with respect to the human bounding box location, and estimate the RSC from the training data. 
Specifically, given a human $h$ and an object $o$, we denote the human bounding box as $(x_h,y_h, w_h, h_h)$ and object bounding box as $(x_o,y_o, w_o, h_o)$, where $(x,y)$ is the top left point and $(w,h)$ is width and height of the bounding box. Inspired by previous work~\cite{gkioxari2018detecting}, we define the RSC as $(\Delta x_{oh}, \Delta y_{oh}, \Delta w_{oh}, \Delta h_{oh})$, where: $\Delta x_{oh}= \frac{x_o-x_h}{w_h}$, $\Delta y_{oh}= \frac{y_o-y_h}{h_h}$, $ \Delta w_{oh}= \log{\frac{w_o}{w_h}}$, and $\Delta h_{oh}= \log{\frac{h_o}{h_h}}$.
We then consider for each interaction, $\Delta x_{oh}, \Delta y_{oh}$ follow a bi-variate Gaussian distribution and $\Delta w_{oh}, \Delta h_{oh}$ follow another bi-variate Gaussian distribution. We estimate essential parameters (mean, co-variance) for these variables using all the training samples for each OA label. We estimate the person bounding boxes in the similar way.   

Using these distributions, we then generate random samples as the human and object bounding boxes to create the spatial features.
As shown in Fig.~\ref{fig:model_architecture_sem_spa} we follow previous works~\cite{ulutan2020vsgnet} to generate the spatial map for each OA label. The spatial map is a $2 \times B \times B$ size binary map where in the first channel the location of the human bounding box is 1 and in the second channel the location of the object bounding box is 1. The rest of the locations in spatial map are zero. Finally we pass through this spatial map to 2 convolution layers followed by a linear projection layer 
to generate the spatial feature. For each OA candidate, $ (y_{o, i}, y_{a, i}) \in {\bf S}_{\text{cand}} $, we compute the spatial features:
\begin{equation}
\vspace{-1.5mm}
\begin{small}
        {\bf f}_{spa}(y_{o, i}, y_{a, i})= \text{Emb}_{spa}(y_{o, i}, y_{a, i})
\end{small} 
\end{equation} 
$\text{Emb}_{spa}$ is the embedding function for the spatial features. 

\vspace{-3mm}

\paragraph{Feature Aggregation.} For each pre-selected OA candidate, $ (y_{o, i}, y_{a, i}) \in {\bf S}_{\text{cand}} $,  we have the semantic feature, ${\bf f}_{sem}(y_{o, i}, y_{a, i})$ and the spatial feature, ${\bf f}_{spa}(y_{o, i}, y_{a, i})$. These features are aggregated as follows: 
\begin{equation}
\begin{small}
        {\bf f}_{agr}(y_{o, i}, y_{a, i})= g_{agr}({\bf f}_{sem}(y_{o, i}, y_{a, i}), {\bf f}_{spa}(y_{o, i}, y_{a, i})) \label{eq:sem_spat_feat_agg}
\end{small} 
\end{equation} 
where $g_{agr}$ is the aggregation function. 
We concatenate the features, $ {\bf f}_{agr}(y_{o, i}, y_{a, i}) $ extracted for all candidates $ \in {\bf S}_{\text{cand}} $ and form the support feature, $ {\bf f}_{supp}({\bf x}) \in \mathbb{R}^{K \times d} $.

\subsection{Query Refiner}\label{subsec:cr}

The query refiner is designed to use the pre-selected OA candidates and support features generated from SFG module to refine the learnt queries that are randomly initialized. Ideally these pre-generated contextual signals should be able to guide the queries to be learnt to attend to more relevant candidates and reduce noisy predictions. To achieve this we cross-attend the learnt queries with support features.   

Specifically, as shown in Fig.~\ref{fig:model_architecture}, the query refiner is built on standard transformer decoder structure. 
The randomly initialized queries ${\bf Q}= \{ {\bf q_i} | {\bf q_i} \in \mathbb{R}^d \}_{i=1}^{N_q}$  first attend to themselves via self-attention. 
Then these queries attend to support features $  {\bf f}_{supp}({\bf x}) $ generated from the SFG (Sec.~\ref{subsec:sem&spa}) through cross-attention. Here, the support features serve as keys and values to the attention architecture. As a result, queries have additional direction to look for correct object-action in the encoded image features. 
In the final HOI decoder, queries attend to the encoded image features. The output of the decoder are the context-aware queries which contain rich cues to detect HOIs. 

\subsection{Training details}\label{sec:train}
To train this network, we apply the same loss function as~\cite{tamura2021qpic} at the outputs of the interaction head, ${\bf b}_{o}$, ${\bf b}_{h}$, ${\bf P}_{HOI}$ and ${\bf P}_{obj}$. The loss calculation is composed of two stages: the bipartite matching stage between predictions and ground truths, and the loss calculation stage for the matched pairs. For the bipartite matching, we follow the training procedure of DETR~\cite{carion2020end} and use the Hungarian algorithm~\cite{kuhn1955hungarian}. Then the loss is calculated on the basis of the matched pairs as follows: 
\vspace{-2mm}
\begin{equation}
 \mathcal{L} =\lambda_1\mathcal{L}_{\text{box}} +\lambda_2\mathcal{L}_{\text{iou}}
 +\lambda_3\mathcal{L}_{\text{obj}}
 +\lambda_4\mathcal{L}_{\text{HOI}},
\end{equation}
where $\mathcal{L}_\text{box}$ and $\mathcal{L}_\text{iou}$ are $l_1$ and $GIoU$ loss applied to both human and object bounding boxes, 
$\mathcal{L}_\text{obj}$ is a cross entropy loss for object prediction, and $\mathcal{L}_\text{HOI}$ is a binary cross entropy loss for interaction prediction. $\lambda_1$, $\lambda_2$, $\lambda_3$ and $\lambda_4$ are hyper-parameters selected following~\cite{tamura2021qpic}.

Additionally we use a binary cross entropy loss for the $ {\bf s} $ output, which corresponds to the image-level (object, action) pair prediction. All these losses are trained in a multi-task setting.

\vspace{-1mm}
\section{Experimental settings}

\paragraph{Dataset \& Metrics.} We evaluate SSRT on the two most popular benchmark datasets: V-COCO~\cite{gupta2015visual} and HICO-DET~\cite{chao2018learning}.
{\bf V-COCO} has 29 interaction classes. Following~\cite{li2019transferable}, we evaluate the performance on 24 interaction classes since 4 interaction classes have no object pair and 1 class has very few samples. This dataset has 2,533 training, 2,867 validation and 4,946 testing images. {\bf HICO-DET}~\cite{chao2018learning} has 600 human-object interaction classes. It consists of 38,117 training and 9,658 test images. 

We report mean average precision (mAP) on the test set for both V-COCO and HICO-DET datasets. A prediction is considered to be correct if the predicted human and object bounding boxes overlap (with IoU greater than 0.5) with the respective GT boxes and the predicted interaction class is correct. We follow the protocol established in~\cite{ulutan2020vsgnet} to evaluate results on the V-COCO dataset. For human actions that do not interact with any object, two evaluation scenarios are considered. Scenario 1 considers a strict evaluation criteria that requires the prediction of a null bounding box with coordinates $[0,0,0,0]$,
Scenario 2 relaxes this condition for such cases by ignoring the predicted bounding box for evaluation. 
We use the protocol from~\cite{chao2018learning} to evaluate on the HICO-DET. The mAP metric is computed in default settings for three categories: Full (all 600 HOI classes), Rare (138 classes that have less than 10 training samples), Non-rare (462 classes that have more than 10 training samples).   

\vspace{-3mm}

\paragraph{Implementation Details.} The architecture design is similar to that of QPIC~\cite{tamura2021qpic}. We use ResNet-50 and ResNet-101 backbones~\cite{he2016deep}. The parameters of the network are initialized with DETR~\cite{carion2020end} trained on the COCO dataset~\cite{tamura2021qpic}. 
Each of the encoder and decoder have 6 layers and 8 heads. The dimension inside the transformer architecture is 256. The total number of queries is 100.   
The initial learning rate of the backbone network is $10^{-5}$, with others $10^{-3}$. The weight decay is $10^{-4}$. The learning rate is dropped at every 65 epochs and we train 150 epochs in total. We use the AdamW~\cite{loshchilov2017decoupled} optimizer and the batch size is 16.

We experiment with the following semantic feature generator: (a) one-hot, (b) GLOVE~\cite{pennington2014glove}, (c) CLIP~\cite{radford2021learning}. In the spatial feature generator, we use a $2 \times 64 \times 64$ dimensional binary spatial map~\cite{ulutan2020vsgnet, gao2018ican}. For the human bounding box location, we select (16, 16) as the fixed top-left point, as in the evaluating HOI datasets interacting human bounding boxes are mostly confined at the top-left corner of the images~\cite{ulutan2020vsgnet}. Both spatial and semantic features are projected to a 256-dimensional space.  

\section{Results} \label{section:results}

In this section, we first compare the performance of our SSRT network with the SOTA methods in Sec.~\ref{sec:res_overall}, followed by an ablation study to validate the design choices in Sec.~\ref{sec:abl}. Finally, we show qualitative analysis in Sec.~\ref{sec:qual}. 

\subsection {Comparison with SOTA}
\label{sec:res_overall}

In Tables~\ref{tab:v-coco} and~\ref{tab:hicodet}, we compare the performance of our SSRT model to the SOTA methods on the V-COCO~\cite{gupta2015visual} and HICO-DET~\cite{chao2018learning} datasets respectively. We group the approaches into one-stage and two-stage. Following the literature, we report numbers of SSRT with both ResNet-50 (R-50) and ResNet-101 (R-101) backbones. Results show that our SSRT has achieved SOTA performance on both datasets with the ResNet-50 backbone, while ResNet-101 can improve the performance further. We outperform all the DETR based solutions (HOI-Trans, ASNet, HOTR and QPIC) on both datasets, and overall we achieve about 10\% improvement on V-COCO and 5\% improvement on HICO-DET comparing to the SOTA. 

\begin{table}[]
\begin{adjustbox}{width=0.90\columnwidth,center}
\centering
\footnotesize
\begin{tabular}{c|lcc}
\hline
Type & Method       & Scenario 1 & Scenario 2       \\ \hline
\parbox[t]{1mm}{\multirow{8}{*}{\rotatebox[origin=c]{90}{Two Stage}}} & VCL~\cite{hou2020visual}  &    48.3 & - \\
& DRG~\cite{gao2020drg} & 51.0 & -\\
& Wang et al.~\cite{wang2020contextual}  & 52.3 & - \\
& FCL~\cite{hou2021detecting} & 52.4 & -  \\
& PD-Net~\cite{zhong2020polysemy} & 52.6 & - \\
& ACP~\cite{kim2020detecting} & 53.0 & -  \\
& FCMNet~\cite{liu2020amplifying} & 53.1 & - \\
& SG2HOI~\cite{he2021exploiting} & 53.3 & -\\
& IDN~\cite{IDN} & 53.3 & 60.3 \\
& GTNet~\cite{iftekhar2021gtnet}                               & 56.2 & 60.1             \\ 
& SABRA~\cite{jin2020towards}                                & 56.6   & -          \\ \hline
\parbox[t]{2mm}{\multirow{10}{*}{\rotatebox[origin=c]{90}{One Stage}}} & UnionDet~\cite{kim2020uniondet} & 47.5 & 56.2 \\
& Wang et al.~\cite{wang2020learning} &  51.0 & - \\ 
& HOI-Trans~\cite{zou2021end} & 52.9 & -\\
& ASNet~\cite{chen2021reformulating}  & 53.9 & - \\
& GGNet~\cite{zhong2021glance}       &   54.7 & -        \\
& HOTR~\cite{kim2021hotr}                                    & 55.2      & 64.4       \\
& DIRV~\cite{fang2020dirv}                                   & 56.1     &  -        \\
& QPIC(R-50)~\cite{tamura2021qpic}                                    & 58.8 & 61.0 \\ 
& QPIC(R-101)~\cite{tamura2021qpic}                                    & 58.3 & 60.7\\ 
\rowcolor{gray!10}
& Ours (R-50) &                             \underline{63.7}  &\underline{65.9}  \\
\rowcolor{gray!10}
& Ours (R-101) &                             \textbf{65.0}  & \textbf{67.1} \\ \hline
\end{tabular}
\end{adjustbox}
\caption{ \small \it Performance comparisons on the  V-COCO~\cite{gupta2015visual} test set. Best result is marked with \textbf{bold} and the second best result is marked with \underline{underline}.}
\label{tab:v-coco}
\end{table}


\begin{table}[]
\begin{adjustbox}{width=0.90\columnwidth,center}
\centering
\footnotesize
\begin{tabular}{c|lccc}
\hline
Type & Method   & Full & Rare & Non-rare       \\ \hline
\parbox[t]{1mm}{\multirow{8}{*}{\rotatebox[origin=c]{90}{Two Stage}}}
&Wang et al.~\cite{wang2020contextual} & 17.57 & 16.85 & 17.78 \\
&  FCMNet~\cite{liu2020amplifying} & 20.41 & 17.34 & 21.56 \\
 & ACP~\cite{kim2020detecting} & 20.59 & 15.92 & 21.98 \\
 & PD-Net~\cite{zhong2020polysemy} & 20.81 & 15.90 & 22.28 \\
  &SG2HOI~\cite{he2021exploiting} & 20.93 & 18.24 & 21.78 \\
 & VCL~\cite{hou2020visual} & 23.63 & 17.21 & 25.55 \\
 & DRG~\cite{gao2020drg} & 24.53 & 19.47 & 26.04 \\
 & SABRA~\cite{jin2020towards}        & 26.09 & 16.29 &    29.02       \\
 & IDN~\cite{IDN} & 26.29 & 22.61 &27.39 \\
 & GTNet~\cite{iftekhar2021gtnet}     & 26.78 & 21.02 &   28.50          \\ 
 & ATL~\cite{hou2021affordance} & 28.53 & 21.64 & 30.59 \\
 & FCL~\cite{hou2021detecting} & 29.12 & 23.67 & 30.75 \\\hline
\parbox[t]{2mm}{\multirow{10}{*}{\rotatebox[origin=c]{90}{One Stage}}} & UnionDet~\cite{kim2020uniondet}  &  17.58 & 11.72 &  19.33 \\
 & Wang et al.~\cite{wang2020learning} & 19.56 & 12.79 & 21.58  \\
 & PPDM~\cite{liao2020ppdm} &  21.73 &  13.78 & 24.10 \\
 & DIRV~\cite{fang2020dirv}  & 21.78 & 16.38 & 23.39              \\
 & HOI-Trans~\cite{zou2021end} & 23.46 & 16.91& 25.41 \\ 
  & PST~\cite{dong2021visual} & 23.93 & 14.98 & 26.60 \\
 & HOTR~\cite{kim2021hotr}       & 25.10 & 17.34 & 27.42             \\
 & ASNet~\cite{chen2021reformulating} & 28.87 & 24.25 & 30.25 \\
  & GGNet~\cite{zhong2021glance}  & 29.17 & 22.13 & 30.84        \\ 
 & QPIC(R-50)~\cite{tamura2021qpic}     &29.07 &21.85 & 31.23 \\
 & QPIC(R-101)~\cite{tamura2021qpic}    &29.90 &23.92 & 31.69 \\
\rowcolor{gray!10}
 & Ours (R-50) &  \underline{30.36} & \textbf{25.42} & \underline{31.83} \\
\rowcolor{gray!10}
 & Ours (R-101)   & \textbf{31.34} & \underline{24.31} & \textbf{33.32}     \\ \hline
\end{tabular}
\end{adjustbox}
\caption{\small \it Performance comparisons on the HICO-DET~\cite{chao2018learning} test set. Best result is marked with \textbf{bold} and the second best result is marked with \underline{underline}.}
\label{tab:hicodet}
\end{table}


\subsection {Ablation Studies}
\label{sec:abl}

In this section, we do ablation for the different design choices of SSRT. We evaluate on V-COCO dataset with the ResNet-50 backbone. For each ablation, we change one parameter, and keep the other parameters at the best setting.

\vspace{-4mm}
\paragraph{Support Feature Generator Module.} In Table~\ref{tab:table1_a}, we explore the benefits of using semantic and spatial features to generate features for the query refiner block. Comparing to the QPIC baseline (Row 1), using semantic features (Row 2) significantly improves the performance by +3.9 points. This demonstrates the effectiveness of using the semantic information to guide the HOI detection. On top of this, we explore two different ways to aggregate the semantic and spatial features: 
(1) concatenation (Row 3); and (2) elementwise multiplication (Row 4). Results show that elementwise multiplication gives the best performance, which we believe is because that multiplication operates as a gating
mechanism that effectively fuses semantic and spatial information, as also observed in other work~\cite{ulutan2020vsgnet,pham2021learning}.

\vspace{-4mm}
\paragraph{Semantic Inputs.} Table~\ref{tab:table1_b} explores different kinds of semantic input that can be encoded as semantic features. For this experiment, all varieties of semantic input are encoded by the CLIP~\cite{radford2021learning} text embedding model. We explore the following types of semantic input: (a) {\it $ action $ only}: using only predicted action category from the encoder. For example, if the OA prediction is $\langle laptop, work \rangle$, we only use the predicted action (i.e., ``work'' here) as the semantic input, (b) {\it object-action (OA)}: Using the previous example, the semantic input here is $\langle laptop, work \rangle$ tuple, (c) {\it semantic retrieval}: In this approach we model the semantic input in a non-parametric fashion. Using a joint visual-semantic embedding network~\cite{radford2021learning}, we retrieve nearest OA semantic tuples based on the visual features of the input. The retrieved candidates are used as semantic input in this case. (d) {\it V-COCO captions}: Since V-COCO is a subset of the COCO dataset~\cite{chen2015microsoft}, we use the corresponding captions as additional input along with the image. In the last row of the table, we also experiment with the oracle setting, where we assume we have access to the ground truth (GT) OA tuple. The strong performance of the oracle model indicates that refining the queries in this manner is an effective direction to guide the network to focus on more relevant candidates. The best performing approach in the non-oracle setting uses OA tuple. There is still a non-trivial gap between it and using the oracle, indicating that there is still room to improve the HOI detection accuracy by further improving the quality of the pre-generated OA tuple candidates.

It is interesting to note that using captions as additional input along with the image does not improve performance. This might be due to the fact that compared to the {\it object-action} candidate, the image captions can be noisy and sometimes too generic for the task (e.g. this photo has a horse etc.). Using only the $ action $ approach achieves a slightly worse performance than using the OA tuple as expected, as the former has no information about the object category.

\begin{table}[t]
\footnotesize
\centering
\begin{minipage}[t]{0.2\textwidth}
\begin{subtable}{\textwidth}
    \begin{tabular}{lc}
        \toprule & mAP    \\ 
        \midrule
        Base (QPIC)             & 58.8 \\
        Base + Sem.       &   62.7 \\ 
        Base + Sem.+ Sp. (concat.) & 62.9 \\ 
        Base + Sem.+ Sp. (multi.) &  \textbf{63.7} \\ 
        \bottomrule
    \end{tabular}
	\caption{\small \it Support Feature Generator Module.}
	\label{tab:table1_a}
\end{subtable}
\vspace{2mm}
\end{minipage} \hfill
\begin{minipage}[t]{0.2\textwidth}
\begin{subtable}{\textwidth}
    \begin{tabular}{lc}
        \toprule & mAP   \\ 
    	\midrule
	    $ action $ only& 63.1  \\ 
		OA & \textbf{63.7} \\ 
		semantic retrieval & 62.7\\ 
    	V-COCO captions & 62.6\\ \midrule
		Oracle HOI GT & 76.1 \\
		\bottomrule
    \end{tabular}
	\caption{\small \it Semantic Inputs.}
	\label{tab:table1_b}
\end{subtable}
\end{minipage}
\begin{minipage}[t]{0.15\textwidth}
\begin{subtable}{\textwidth}
    \begin{tabular}{lccc}
        \toprule
        \#   & Prec. & Rec. & mAP  \\ 
    	\midrule
    	1  & 85.3 & 23.4 & 62.9\\
	    2  & 75.1 & 41.2 & 63.3 \\ 
		4  & 62.5 & 68.6 & \textbf{63.7} \\ 
		8  & 37.1 & 81.5 & 63.4\\ 
		13 & 25.1 & 89.6 & 62.8\\
		\bottomrule
    \end{tabular}
	\caption{\small \it Numbers of HOI predictions selected as semantics. Prec. is precision and Rec. is recall.}
	\label{tab:table1_bc}
\end{subtable}
\end{minipage} \hfill
\begin{minipage}[t]{0.25\textwidth}
\begin{subtable}{\textwidth}
    \begin{tabular}{lc}
        \toprule
    	& mAP   \\ 
    	\midrule
		One-hot vector & 63.0    \\ 
		GLOVE embedding~\cite{pennington2014glove} & 63.1 \\ 
		CLIP text embedding~\cite{radford2021learning} & \textbf{63.7}\\
        \bottomrule
    \end{tabular}
	\caption{\small \it Semantic embeddings.}
	\label{tab:table1_c}
\end{subtable}
\end{minipage}
\begin{minipage}[t]{0.5\textwidth}
\begin{subtable}{\textwidth}
\centering
\begin{tabular}{lc}
    \toprule &   mAP   \\ 
	\midrule
	Multi-variate parameter & 62.3 \\
	Bi-variate parameter & 62.0 \\ 
	Multi-variate spatial map & 63.0    \\ 
	Bi-variate spatial map & \textbf{63.7} \\ 
    \bottomrule
\end{tabular}
\caption{\small \it Spatial feature designs.}
\label{tab:table1_d}
\end{subtable}
\end{minipage}
\caption{\small \it Design choices on the semantic and spatial features.}
\vspace{-5mm}
\label{tab:table1}
\end{table}

 
\vspace{-4mm}
\paragraph{Number of Predictions as Semantics.} We then ablate the different number of OA predictions candidate selected as the semantic inputs in Table~\ref{tab:table1_bc}. We test with using topK, 
where K=1, 2, 4, 8 and 13 HOI predictions as semantics. We stop at K=13 because the maximum number of HOI ground truths for V-COCO in each image is 13. To better understand the results, we not only list the final mAP metric, but also add the precision and recall for the prediction in the table. Results show that K=4 gives the best performance, and the performance gradually decreases when moving towards either direction (K=1 and K=13). As expected, K=1 gives the highest precision for prediction and K=13 gives the highest recall. But the optimal performance point (K=4) is in the middle, indicating that the trade-off between precision and recall of the prediction is important. Low recall corresponds to using lesser information for the query refiner block to produce representative enough queries for the decoder. Low precision affects the quality of input to the refiner block with increasing noise. 
 
\vspace{-4mm}
\paragraph{Semantic Embeddings.} 
We then evaluate different embedding methods in Table~\ref{tab:table1_c}. We test with (1) a one-hot vector from the prediction; (2) the GLOVE~\cite{pennington2014glove} encoder,  
and (3) the CLIP~\cite{radford2021learning} text encoder. Results show that all embeddings achieve good performance, while CLIP achieves the best. This may due to the fact that CLIP encoder is learnt from large-scale image text-pairs and hence generates a stronger semantic embedding for the HOI task than others. One-hot results also give good performance, indicating that using the pre-selected OA candidates itself can still provide guidance to refine the queries.

\begin{figure}
\begin{center}
\includegraphics[width=1\linewidth]{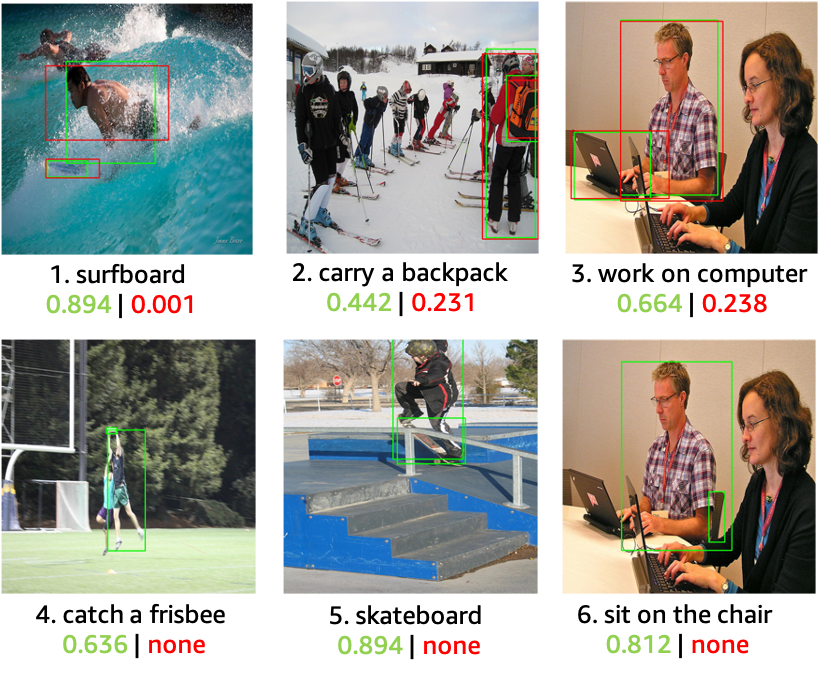}
\end{center}
\vspace{-0.6cm}
  \caption{\small \it Qualitative results of SSRT compared to QPIC. For each image, the detection outputs of SSRT are marked in green while the outputs of QPIC are marked in red. The prediction scores are presented in the captions. If no matched bounding box pairs are detected then the score is marked as none. We observe that SSRT improves over QPIC in mainly two aspects: (1) increasing the confidence scores of the interaction predictions (sample 1-3); and (2) successfully detecting the person, object and interactions that are completely missed in QPIC (sample 4-6).}
\label{fig:err}
\end{figure}

\begin{figure*}
        \centering
        \begin{subfigure}[b]{0.475\textwidth}
            \centering
            \includegraphics[width=\textwidth]{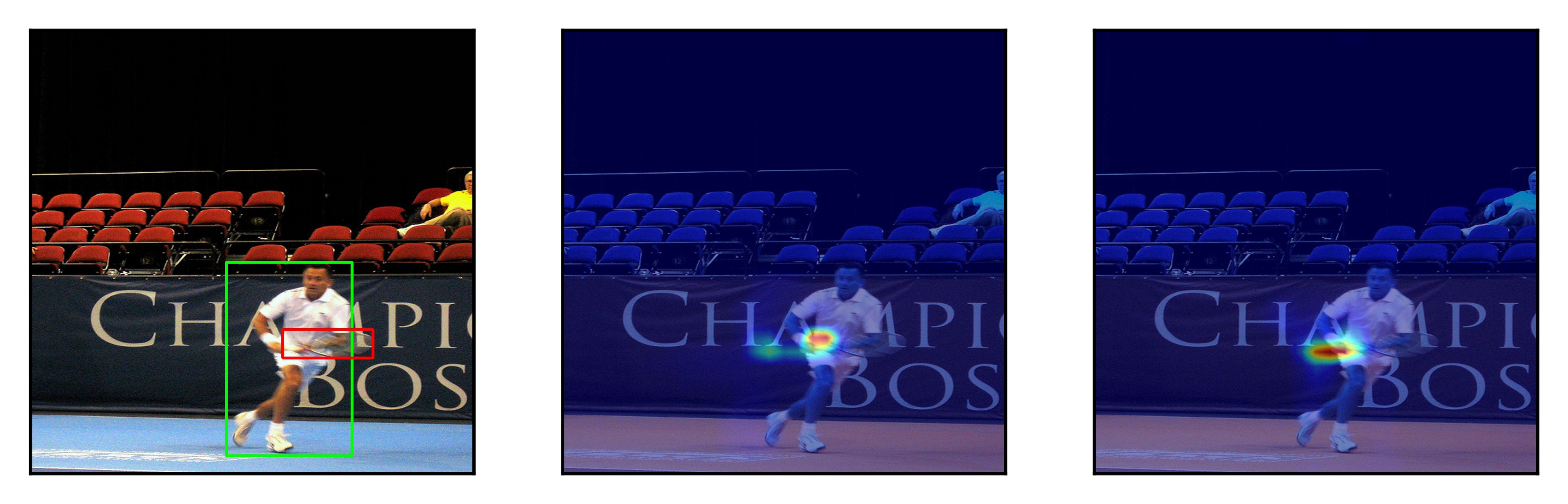}
            \caption[]%
            {Hold a tennis racket. Scores: \textcolor{blue}{SSRT: 0.856} $|$ \textcolor{red}{QPIC: 0.001}.}
            \label{fig:att1}
        \end{subfigure}
        \hfill
        \begin{subfigure}[b]{0.475\textwidth}  
            \centering 
            \includegraphics[width=\textwidth]{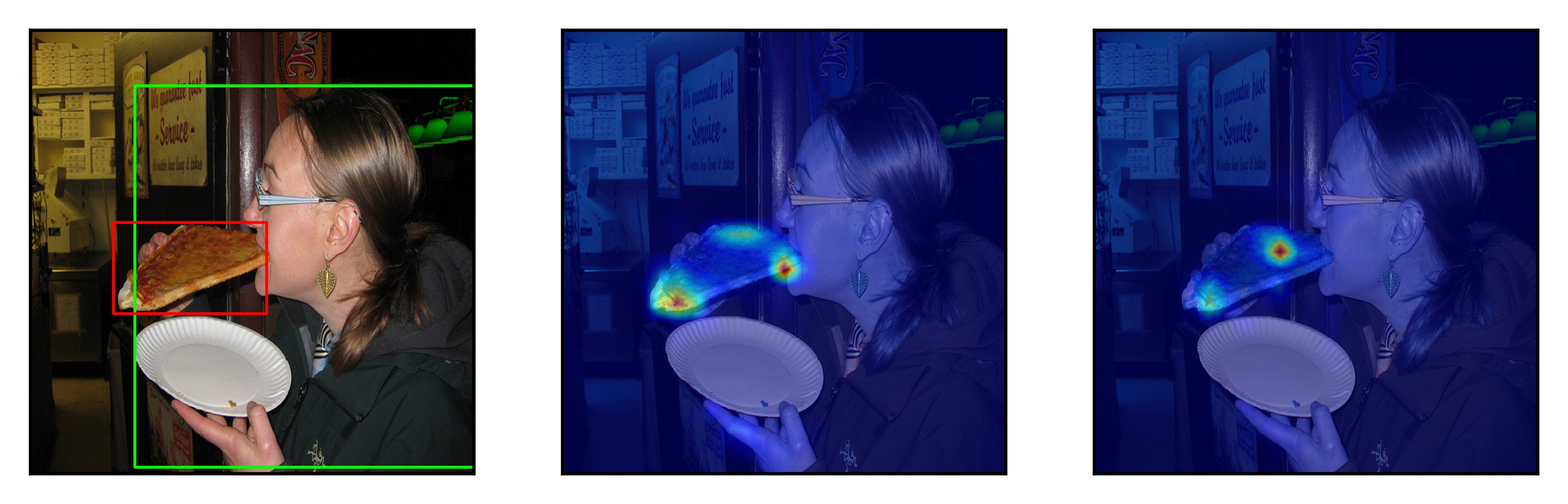}
            \caption[]%
            {Eat a pizza. Scores: \textcolor{blue}{SSRT:  0.761} $|$ \textcolor{red}{QPIC: 0.512}. }    
            \label{fig:att2}
        \end{subfigure}
        \begin{subfigure}[b]{0.475\textwidth}   
            \centering 
            \includegraphics[width=\textwidth]{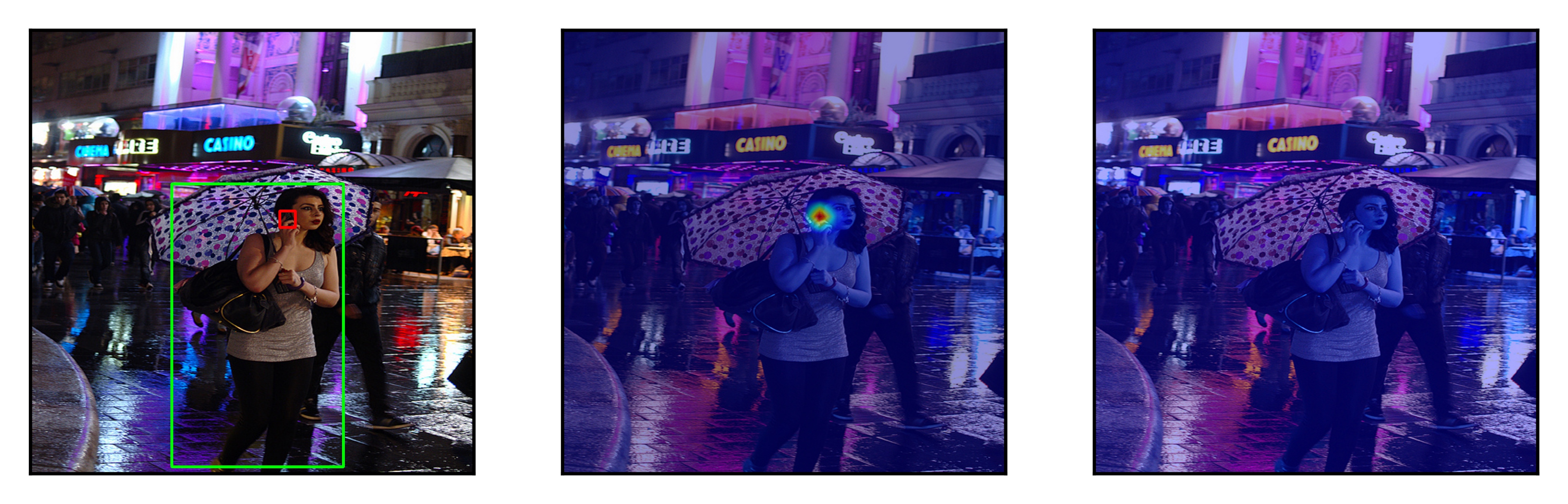}
            \caption[]%
            {Talk on the phone. Scores: \textcolor{blue}{SSRT:  0.612} $|$ \textcolor{red}{QPIC: none}.}    
            \label{fig:att3}
        \end{subfigure}
        \hfill
        \begin{subfigure}[b]{0.475\textwidth}   
            \centering 
            \includegraphics[width=\textwidth]{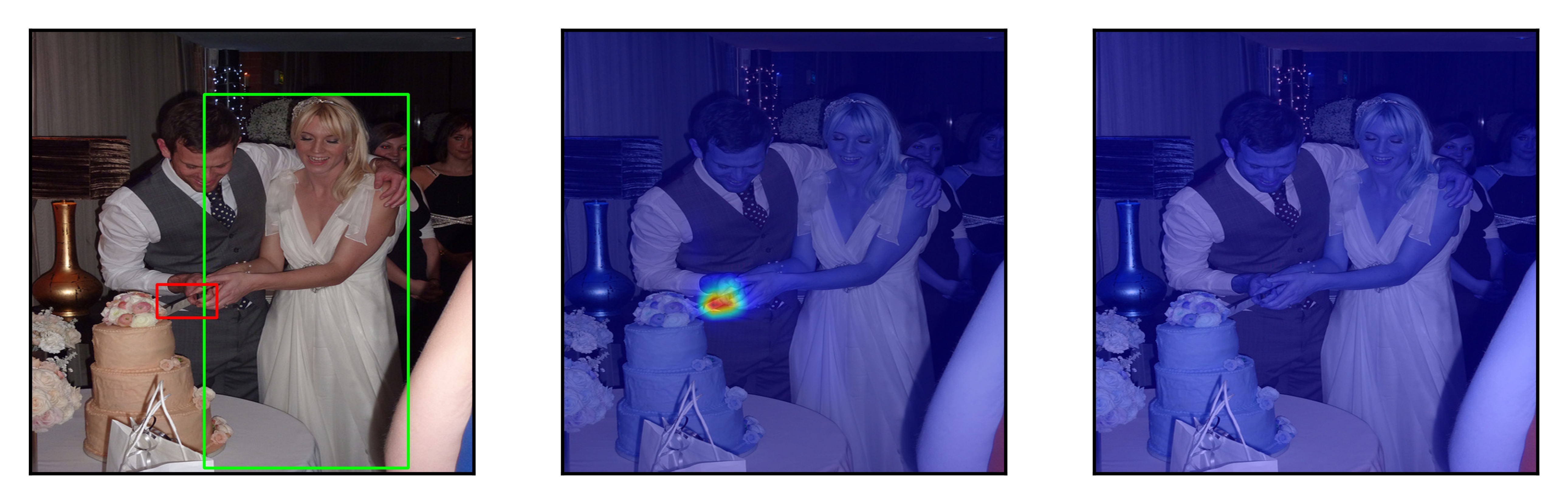}
            \caption[]%
            {Cut with knife. Scores: \textcolor{blue}{SSRT:  0.757} $|$ \textcolor{red}{QPIC: none}.}    
            \label{fig:att4}
            \vspace{-1mm}
        \end{subfigure}
        \caption[]
        {\small \it Visualization of the attention. We extract the attention map from the last layer of the decoder. In each sub-figure, from the left to the right are (1) the original image with the ground truth; (2) the attention map of our SSRT, and (3) the attention map of QPIC.} 
        \label{fig:att}
        \vspace{-0.3cm}
    \end{figure*}

\vspace{-4mm}
\paragraph {Spatial Feature Designs.} We evaluate the performance of different spatial feature designs in Table~\ref{tab:table1_d}. For the relative spatial configuration (RSC) introduced in Sec.~\ref{subsec:sem&spa}, we consider $(\Delta x_{oh}, \Delta y_{oh}, \Delta w_{oh}, \Delta h_{oh})$ to either follow a multi-variate distribution or follow two bi-variate distributions for $(\Delta x_{oh}, \Delta y_{oh})$ and $(\Delta w_{oh}, \Delta h_{oh})$. With each distribution, we explore two types of features: (1) only using parameters of distributions as features (Row 1 and 2). Specifically, for multi-variate distribution we use mean, variance and co-variance among all combinations of $\Delta x_{oh}, \Delta y_{oh}, \Delta w_{oh}, \Delta h_{oh}$ as the feature, and for bi-variate distribution we use the mean and variance of all $\Delta x_{oh}, \Delta y_{oh}, \Delta w_{oh}, \Delta h_{oh}$, plus only co-variance of $(\Delta x_{oh}, \Delta y_{oh})$ and of $(\Delta w_{oh}, \Delta h_{oh})$ as the feature. We concatenate them with semantic features as multiplication is not an option here; (2) we generate random samples from the distribution and then create the spatial map (Row 3 and 4) as introduced in Sec.~\ref{subsec:sem&spa}. From Table~\ref{tab:table1_d} we can see that using spatial map always outperforms directly using parameters as features, which we believe is due to that spatial maps have a much higher dimension (2 x 64 x 64) than direct parameters (14 or 17) that can learn richer spatial configurations. In addition, the bi-variate distribution generated spatial map outperforms the multi-variate one.

\vspace{-4mm}
\paragraph {Increased number of parameters of QPIC.} Our design improves over QPIC by adding two novel modules (SFG and QR). To validate that SSRT's performance gain is from its design, rather than from its additional model capacity, we now experiment by increasing the number of parameter in QPIC's FFN to match those our approach (49.8M). Interestingly, QPIC's performance drops from 58.8 mAP to 57.9 mAP when its parameters are increased from 41.1M to 49.8M, likely due to overfitting. This clearly shows that SSRT performance (63.7 mAP) comes from our design.

\vspace{-4mm}
\paragraph {Different ways of incorporating the information.} Finally, we test three different ways of incorporating the semantic/ spatial information: (i) between the backbone and the encoder, (ii) as the input to the decoder, instead of using additional cross-attentions, (iii) to initialize the queries. None of these successfully matched SSRT's performance and didn't even improved over the performance of our QPIC baseline. This shows the importance of using such information properly. Upon analyzing these unsuccessful designs, we found that they were sensitive to the accuracy of the object-action (OA) selection and only when the ground truth OAs were used, their performance was better than QPIC. In contrast, SSRT is more robust to changes in OA selections, likely because it only uses the information as support features through additional cross-attention.

\subsection {Qualitative Results}
\label{sec:qual}

We show qualitative results of our SSRT and compare it with the baseline (QPIC). Fig.~\ref{fig:err} shows results of examples selected from different interaction classes. We find that SSRT improves over QPIC mainly in two categories: (1) increasing the confidence scores of the action predictions (case 1-3); and (2) successfully detecting the person, object and actions that are completely missed (no bounding box output matches with GT) in QPIC (case 4-6). These improvement comes across different scenarios including: (1) small or nearly invisible objects (Sample 1, 4, 5, 6); (2) complex scenes (Sample 2); (3) multiple HOI predictions (Sample 3 and 6). 

To further understand the network behavior, we compare the attention maps from SSRT and QPIC in Fig.~\ref{fig:att}. Specifically, we extract the visual attention maps of the query that predicts the marked person and object bounding boxes from the last layer of the decoder. In Fig.~\ref{fig:att1}, both QPIC and SSRT can localize the person and the object, but QPIC fails to predict the action with a high confidence while SSRT does. Looking at the attention map we can see the attention from QPIC is on the roughly correct region but very coarse and noisy, while it from SSRT is much more refined and focused on the area of the interaction (the hand). Similarly in Fig.~\ref{fig:att2}, SSRT achieves higher confidence than QPIC, as the attention is more refined and focused on the interaction area (the mouth and the hand), while QPIC just focuses on the pizza. 
For images in Fig.~\ref{fig:att3} and Fig.~\ref{fig:att4}, QPIC completely misses the prediction while SSRT detects the full correct HOI. We see from the attention map that SSRT is able to attend to the right area while QPIC fails. 
Overall we see that SSRT has more refined and sharper attention, and is able to focus on small objects in complex scenes.

\section{Conclusion}
We proposed SSRT, a one-stage semantic and spatial refined transformer for detecting HOIs. SSRT generates semantic and spatial features based on pre-selected human-object prediction candidates and leverages them to not only enrich the context but also guide the queries to attend to more related predictions. 
SSRT achieved SOTA performance on both V-COCO and HICO-DET datasets, demonstrating the effectiveness of our solution.  

\vspace{-0.5cm}
\paragraph{Limitation.} Our approach requires fully-supervised HOI annotations for training, which are however extremely expensive to collect. In the future, it is important to explore novel HOI solutions that can learn from limited annotations and with less supervision.

\vspace{-0.5cm}
\paragraph{Licenses.} We use the following datasets: V-COCO (CC BY 4.0 license), HICO-DET~\cite{hicolic} and code packages: QPIC~\cite{qpic}, CLIP~\cite{clip}, GLOVE\cite{glove}.

{\small
\bibliographystyle{ieee_fullname}
\bibliography{egbib}
}

\end{document}


\title{What to Look at and Where: Semantic and Spatial Refined Transformer \\ for Detecting Human-Object Interactions \\ \textit{Supplementary Material}}


\author{A S M Iftekhar\thanks{Equal contribution.} \thanks{Work done during an internship at Amazon.}, Hao Chen\footnotemark[1] \textsuperscript{\rm $\ddagger$}, Kaustav Kundu\textsuperscript{\rm $\ddagger$}, Xinyu Li\textsuperscript{\rm $\ddagger$}, Joseph Tighe\textsuperscript{\rm $\ddagger$}, and Davide Modolo\textsuperscript{\rm $\ddagger$}\\
\textsuperscript{\rm $\ddagger$}AWS AI Labs; \textsuperscript{\rm $\dagger$}University of California, Santa Barbara\\
{\tt\small \{hxen,kaustavk,-,tighej,dmodolo\}@amazon.com; iftekhar@ucsb.edu}
}

\maketitle

In this supplementary material, we first analyze the per interaction category results of our SSRT model, then we provide additional qualitative results.

\section{Per Interaction Category Results}
\label{sec:per_cls}
In Table~\ref{tab:category}, we compare the per interaction category results of our SSRT (with ResNet-50 as the backbone) to QPIC~\cite{tamura2021qpic} on V-COCO~\cite{gupta2015visual} dataset under the Scenario 1 setting. Results show that our SSRT improves the performance of all categories over the QPIC without any regression. Closely looking at the table, the top improvements are from: (1) read-obj (+16.04); (2) drink-instr (+11.43); (3) talk-on-phone-instr (+11.39); (4) cut-instr (+9.27); and (5) eat-obj (+7.81). Most of them are interactions with small objects (e.g., books, bottles, phones, scissors, knifes, sandwiches, etc.). While QPIC has low performance in detecting and predicting those categories, SSRT, with enriched semantic and spatial features and refined queries, is able to better focus on these small objects and corresponding interactions, hence improve the performance significantly. On the other hand, we also look into categories that are with smallest improvements. They are: (1) look-obj (+0.57); (2) lay-instr (+0.82); (3) skateboard-instr (+ 0.99); and (4) ride-instr (+1.06). These are interactions either with abstract concepts (e.g., look-obj), or with relatively large objects (e.g., beds, horses, motorcycles, skateboards). Though SSRT still improves performance on these categories, the improvement comparing to QPIC is not as significant as on categories with small and hardly visible objects. As a future work we will explore how to further improve the performance on these categories.

\begin{table}[t]
\centering
\begin{tabular}{|l|r|r|}
\hline
Interaction Category           & QPIC  & SSRT  \\ \hline
hold-obj:       & 50.61   & \textbf{55.45}  \\
sit-instr:              & 51.98                                                                                                                     & \textbf{56.14}                   \\
ride-instr:             & 67.37                                                                                                                    & \textbf{68.43}                   \\
look-obj:               & 47.30                                                                                                          & \textbf{47.87}                            \\
hit-instr:              & 74.66                                                                                                                     & \textbf{79.02}                    \\
hit-obj:                & 66.52                                                                                                                  & \textbf{72.22}                   \\
eat-obj:                & 58.80                                                                                                   & \textbf{66.61}                            \\
eat-instr:              & 72.64                                                                                                                & \textbf{76.06}                   \\
jump-instr:             & 77.81                                                                                                         &  \textbf{80.41}                            \\
lay-instr:              & 54.62                                                                                           & 
\textbf{55.44}                            \\
talk-on-phone-instr:    & 40.26                                                                                                           & \textbf{51.65}                            \\
carry-obj:              & 41.45                                                                                                                     & \textbf{44.53}                   \\
throw-obj:              & 53.21                                                                                                           & \textbf{54.78}                            \\
catch-obj:              & 54.47                                                                                                                  & \textbf{57.52}                   \\
cut-instr:              & 38.07                                                                                                                     & \textbf{47.34}                   \\
cut-obj:                & 58.15                                                                      & \textbf{63.78}                            \\
work-on-computer-instr: & 68.18                                                                                               & \textbf{73.05}                            \\
ski-instr:              & 49.31                                                                                                         & \textbf{52.59}                            \\
surf-instr:             & 70.4                                                                                                                    & \textbf{75.25}                   \\
skateboard-instr:       & 84.43                                                                                                                    & \textbf{85.42}                    \\
drink-instr:            & 44.26                                                                                                                      & \textbf{55.69}                   \\
kick-obj:               &  81.93                                                                                                          & \textbf{84.14}                            \\
read-obj:               & 35.76                                                                                                  & \textbf{51.80}                            \\
snowboard-instr:        & 68.68                                                                                                                 & \textbf{74.27}                   \\ \hline
mAP                     &  58.79                                                                                                                   &  \textbf{63.73}                    \\ \hline
\end{tabular}
\caption{Our network's category-wise performance compare to QPIC~\cite{tamura2021qpic}. The best performance in each row are marked with \textbf{bold}. Here, ``instr'' means instrument and ``obj'' means object~\cite{gupta2015visual}.}
\label{tab:category}
\end{table}

\section{Additional Qualitative Results}

\begin{figure*}[t]
\centering
        \begin{subfigure}[b]{1\textwidth}
            \centering
            \includegraphics[width=\textwidth]{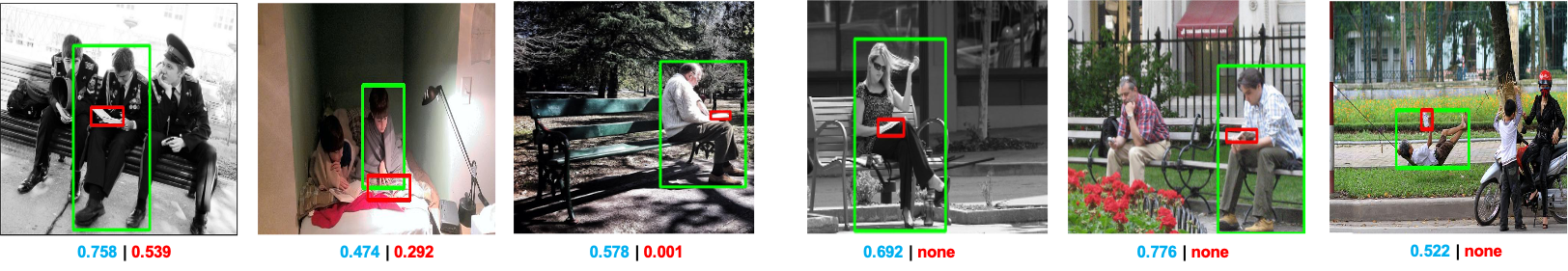}
            \caption[]%
            {Read Object.}
            \label{fig:read}
        \end{subfigure}
        \vskip\baselineskip
        \begin{subfigure}[b]{1\textwidth}
            \centering
            \includegraphics[width=\textwidth]{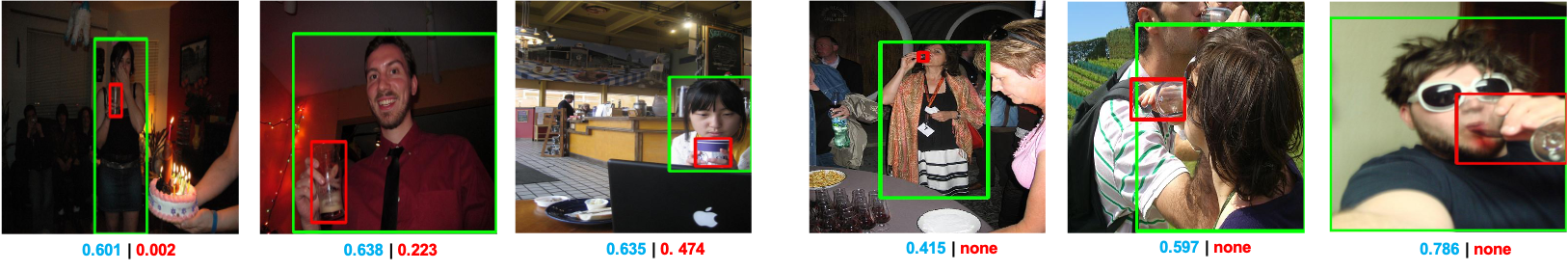}
            \caption[]%
            {Drink Instrument.}
            \label{fig:drink}
        \end{subfigure}
        \vskip\baselineskip
        \begin{subfigure}[b]{1\textwidth}
            \centering
            \includegraphics[width=\textwidth]{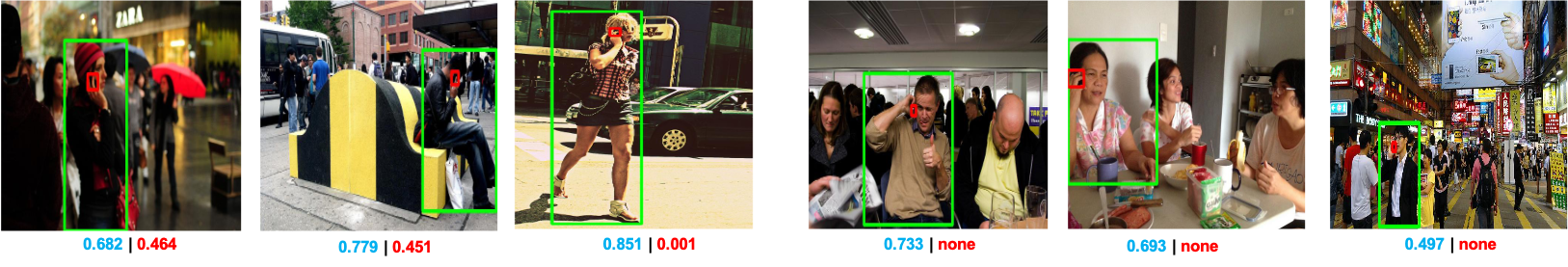}
            \caption[]%
            {Talk on Phone Instrument.}
            \label{fig:talk}
        \end{subfigure}
        \vskip\baselineskip
        \begin{subfigure}[b]{1\textwidth}
            \centering
            \includegraphics[width=\textwidth]{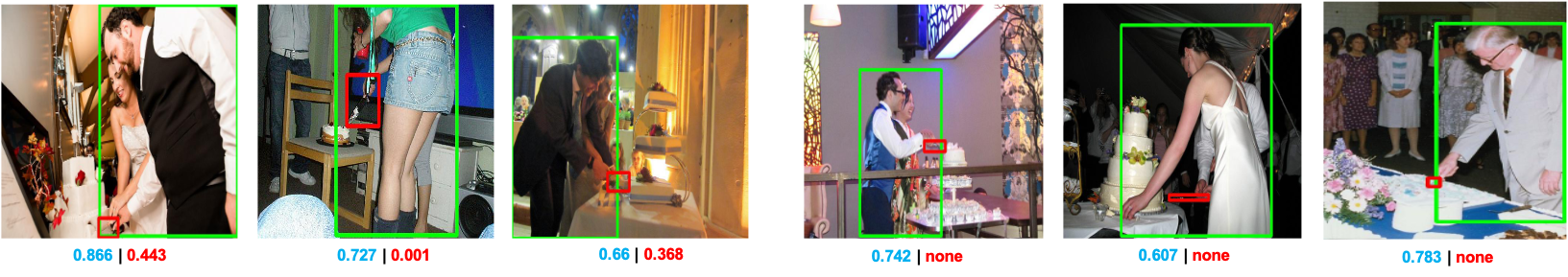}
            \caption[]%
            {Cut Instrument.}
            \label{fig:cut}
        \end{subfigure}
        \vskip\baselineskip
        \begin{subfigure}[b]{1\textwidth}
            \centering
            \includegraphics[width=\textwidth]{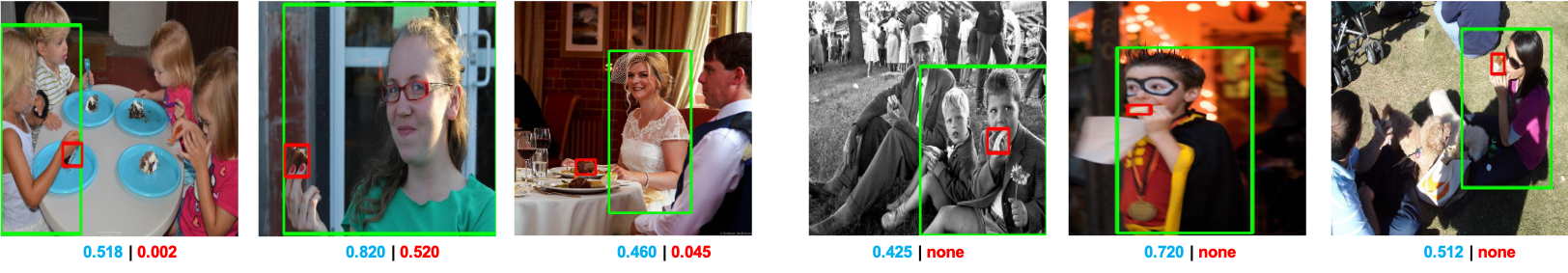}
            \caption[]%
            {Eat Object.}
            \label{fig:eat}
        \end{subfigure}
                \caption[]{Qualitative results for top-5 interaction categories with the biggest improvements from SSRT compared to QPIC. For each image, the predicting score of SSRT is marked in blue while the score of QPIC is marked in red. If no matched bounding box pairs are detected then the result is marked as none.}
\label{suppl:top5}
\end{figure*}

\begin{figure*}[t]
        \centering
        \begin{subfigure}[b]{0.475\textwidth}
            \centering
            \includegraphics[width=\textwidth]{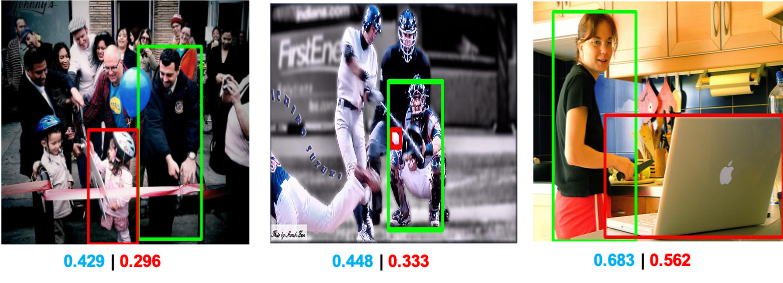}
            \caption[]%
            {Look Object.}
            \label{fig:look}
        \end{subfigure}
        \hfill
        \begin{subfigure}[b]{0.475\textwidth}  
            \centering 
            \includegraphics[width=\textwidth]{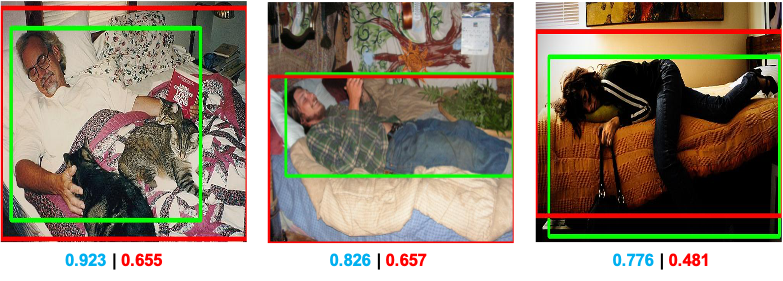}
            \caption[]%
            {Lay Instrument. }    
            \label{fig:lay}
        \end{subfigure}
        \vskip\baselineskip
        \begin{subfigure}[b]{0.475\textwidth}   
            \centering 
            \includegraphics[width=\textwidth]{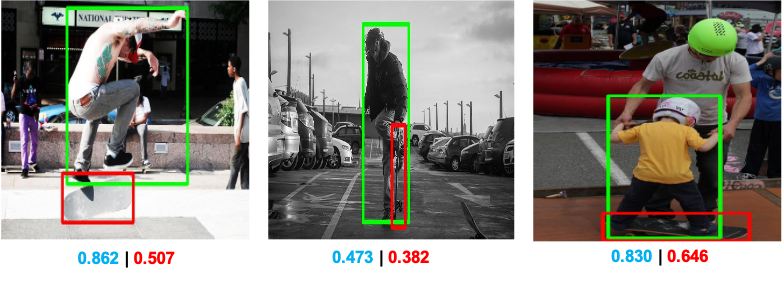}
            \caption[]%
            {Skateboard Instrument.}   
            \label{fig:skateboard}
        \end{subfigure}
        \hfill
        \begin{subfigure}[b]{0.475\textwidth}   
            \centering 
            \includegraphics[width=\textwidth]{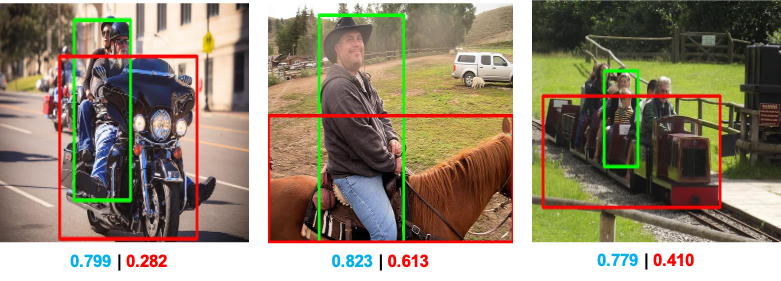}
            \caption[]%
            {Ride Instrument.}    
            \label{fig:ride}
        \end{subfigure}
        \caption[]
        {Qualitative results for interaction categories with the least improvements from SSRT compared to QPIC. For each image, the predicting score of SSRT is marked in blue while the score of QPIC is marked in red.}
        \label{suppl:least}
    \end{figure*}

Fig.~\ref{suppl:top5} and Fig.~\ref{suppl:least} include additional qualitative results of SSRT (with ResNet-50 as the backbone) and the comparison of it with QPIC. Specifically, in Fig.~\ref{suppl:top5}, we show samples from the top-5 interaction categories with the largest performance improvement from SSRT, as introduced in Sec.~\ref{sec:per_cls}. We observe that most of samples from these top-5 categories are persons interacting with small or hardly visible objects in some complex scenes. As mentioned in the main paper, SSRT improves over QPIC mainly in two types: (1) increasing the confidence scores of the action predictions - which is shown in the first three samples of each row in Fig.~\ref{suppl:top5}; and (2) successfully detecting the person, object and actions that are completely missed (no bounding box output matches with GT) in QPIC - which is shown in the last three samples of each row in Fig.~\ref{suppl:top5}).  

Fig.~\ref{suppl:least} shows samples from the interaction categories that are with least improvement from SSRT, as introduced in Sec.~\ref{sec:per_cls}. Note that, though the improvements on these categories are not as big as the top-5 categories, SSRT still performs better than QPIC on all of these categories. We notice that samples from these classes are either with abstract interaction concepts, or contain interactions with relatively larger objects comparing to those in Fig.~\ref{suppl:top5}.

\begin{figure*}[t]
        \centering
        \begin{subfigure}[b]{0.475\textwidth}
            \centering
            \includegraphics[width=\textwidth]{Images/att1_small.jpg}
            \caption[]%
            {Hold a tennis racket. Scores: \textcolor{blue}{SSRT: 0.856} $|$ \textcolor{red}{QPIC: 0.001}.}
            \label{fig:att1}
        \end{subfigure}
        \hfill
        \begin{subfigure}[b]{0.475\textwidth}  
            \centering 
            \includegraphics[width=\textwidth]{Images/att5_small.jpg}
            \caption[]%
            {Eat a pizza. Scores: \textcolor{blue}{SSRT:  0.761} $|$ \textcolor{red}{QPIC: 0.512}. }    
            \label{fig:att5}
        \end{subfigure}
        \vskip\baselineskip
        \begin{subfigure}[b]{0.475\textwidth}   
            \centering 
            \includegraphics[width=\textwidth]{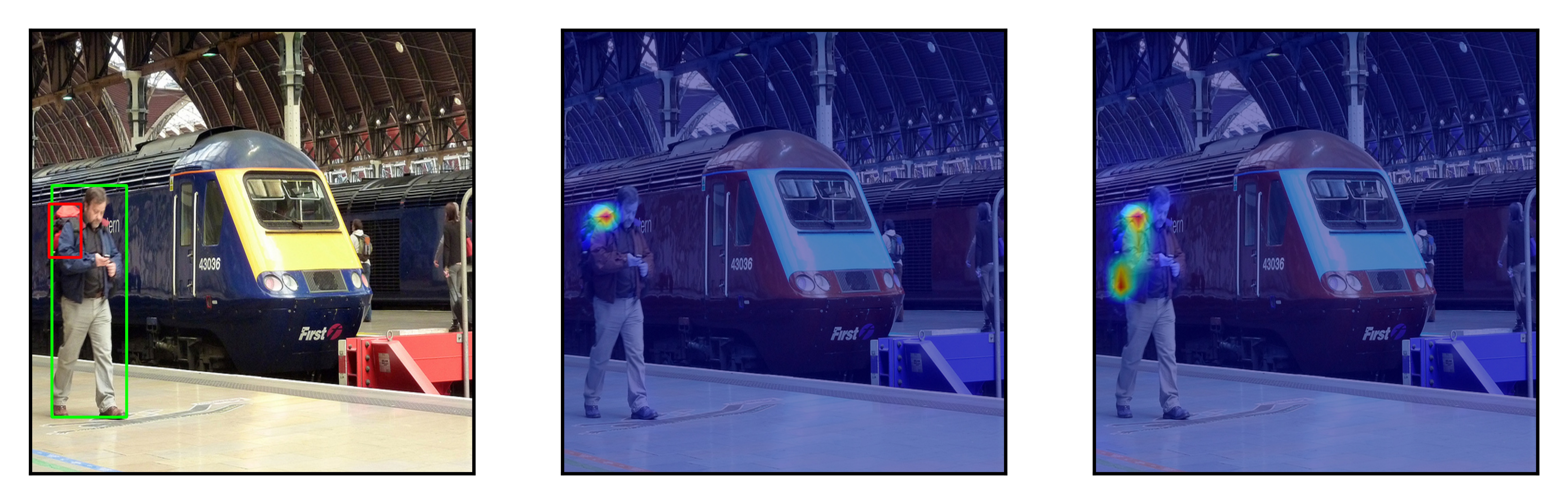}
            \caption[]%
            {Carry a backpack. Scores: \textcolor{blue}{SSRT: 0.588} $|$ \textcolor{red}{QPIC:0.001}.}   
            \label{fig:att2}
        \end{subfigure}
        \hfill
        \begin{subfigure}[b]{0.475\textwidth}   
            \centering 
            \includegraphics[width=\textwidth]{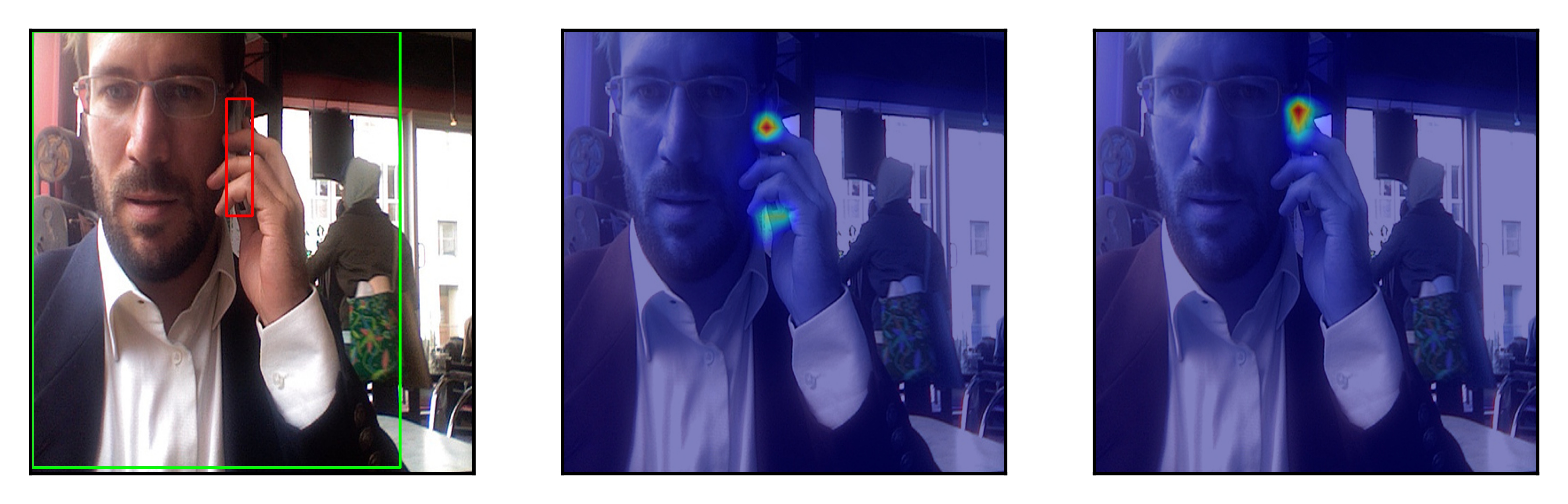}
            \caption[]%
            {Talk on the phone. Scores: \textcolor{blue}{SSRT:  0.880} $|$ \textcolor{red}{QPIC: 0.001}.}    
            \label{fig:att6}
        \end{subfigure}
        \vskip\baselineskip
        \begin{subfigure}[b]{0.475\textwidth}   
            \centering 
            \includegraphics[width=\textwidth]{Images/att3_small.jpg}
            \caption[]%
            {Talk on the phone. Scores: \textcolor{blue}{SSRT:  0.612} $|$ \textcolor{red}{QPIC: none}.}    
            \label{fig:att3}
        \end{subfigure}
        \hfill
        \begin{subfigure}[b]{0.475\textwidth}   
            \centering 
            \includegraphics[width=\textwidth]{Images/att4_small.jpg}
            \caption[]%
            {Cut with knife. Scores: \textcolor{blue}{SSRT:  0.757} $|$ \textcolor{red}{QPIC: none}.}    
            \label{fig:att4}
        \end{subfigure}
        \vskip\baselineskip
        \begin{subfigure}[b]{0.475\textwidth}   
            \centering 
            \includegraphics[width=\textwidth]{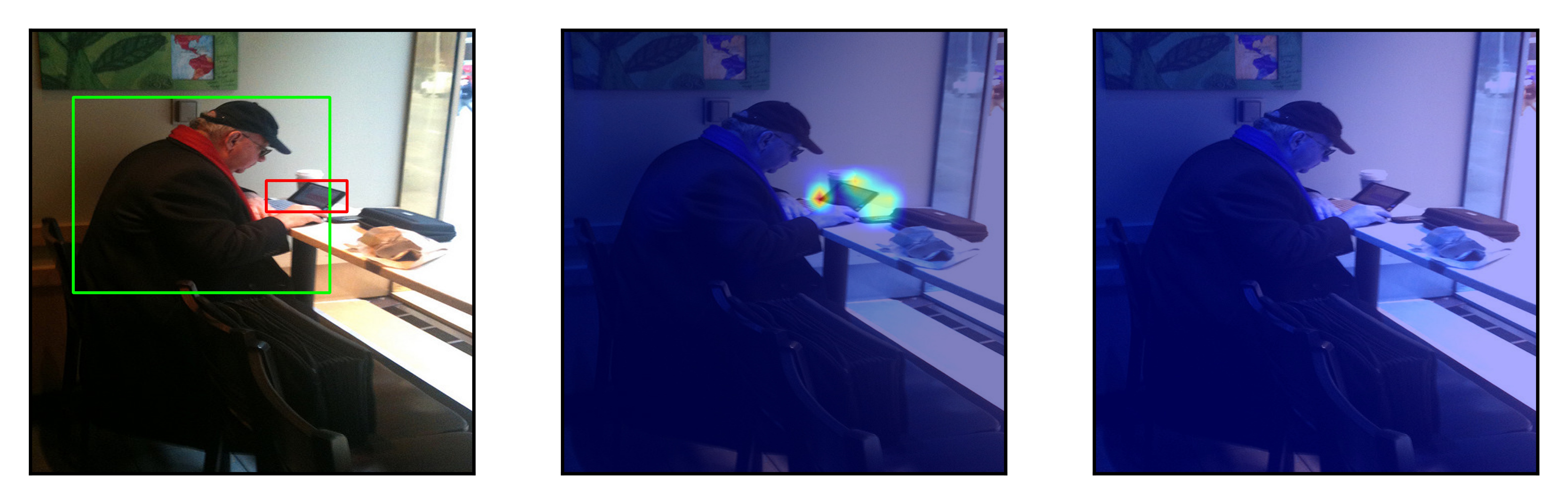}
            \caption[]%
            {Work on the computer. Scores: \textcolor{blue}{SSRT:  0.761} $|$ \textcolor{red}{QPIC: none}.}    
            \label{fig:att7}
        \end{subfigure}
        \hfill
        \begin{subfigure}[b]{0.475\textwidth}   
            \centering 
            \includegraphics[width=\textwidth]{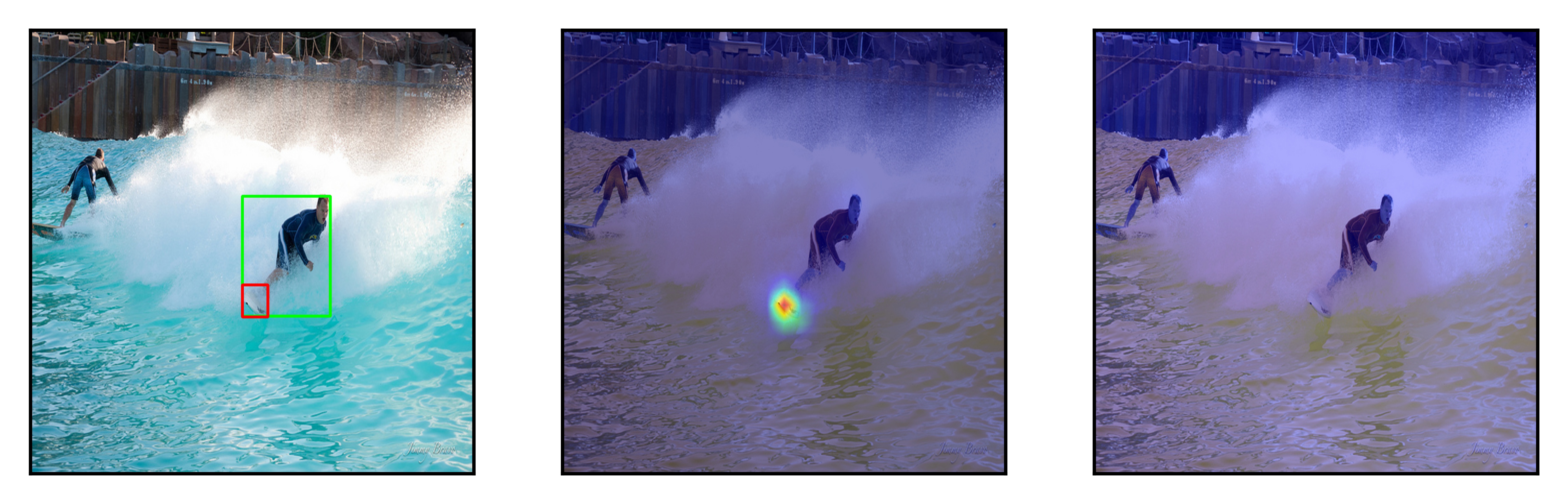}
            \caption[]%
            {Play a surfboard. Scores: \textcolor{blue}{SSRT:  0.873} $|$ \textcolor{red}{QPIC: none}.}    
            \label{fig:att8}
        \end{subfigure}
        \caption[]
        {Visualization of the attention. We extract the attention map from the last layer of the decoder. In each sub-figure, from the left to the right are (1) the original image with the ground truth; (2) the attention map of our SSRT, and (3) the attention map of QPIC.}
        \label{suppl:att}
    \end{figure*}

Fig.~\ref{suppl:att} shows some more visualizations of the attention maps. Recall that the attention map is of the query that predicts the marked person and object bounding boxes, generated from the last layer of the decoder.  For samples in Fig.~\ref{fig:att1} - Fig.~\ref{fig:att6}, both QPIC and SSRT can localize the persons and the objects, but QPIC fails to predict the actions with high confidences while SSRT does. The attention maps clearly show that attentions from SSRT are more refined and focused on the area of the interaction, while attentions from QPIC are on the roughly correct regions but very coarse and noisy. For samples in Fig.~\ref{fig:att3} - Fig.~\ref{fig:att8}, QPIC completely fails in even detecting the object and interaction locations, while SSRT is able to detect to the right area. Most of these samples are persons interacting with small or hardly visible objects in some complex scenes, indicating that SSRT is especially better at handling such scenarios than QPIC.

{\small
\bibliographystyle{ieee_fullname}
\bibliography{egbib}
}